# Thermo-responsive closing and reopening artificial Venus Flytrap utilizing shape memory elastomers

Shun Yoshida[1], Qingchuan Song[2,3][0000-0003-2548-8009], Bastian E. Rapp[2,3,4][0000-0002-3955-0291], Thomas Speck[1,2,4][0000-0002-2245-2636], Falk J. Tauber [1,2,4][0000-0001-7225-1472]

[1]Plant Biomechanics Group (PBG) Freiburg, Botanic Garden Freiburg, University of Freiburg, Germany
[2]Cluster of Excellence livMatS @ FIT – Freiburg Center for Interactive Materials and Bioinspired Technologies, University of Freiburg, Germany
[3]Laboratory of Process Technology, NeptunLab, Department of Microsystem Engineering (IMTEK), University of Freiburg, Germany
[4]Freiburg Materials Research Center (FMF), University of Freiburg, Germany
`falk.tauber@biologie.uni-freiburg.de`

**Abstract.** Despite their often perceived static and slow nature, some plants can move faster than the blink of an eye. The rapid snap closure motion of the Venus flytrap (*Dionaea muscipula*) has long captivated the interest of researchers and engineers alike, serving as a model for plant-inspired soft machines and robots. The translation of the fast snapping closure has inspired the development of various artificial Venus flytrap (AVF) systems. However, translating both the closing and reopening motion of *D. muscipula* into an autonomous plant inspired soft machine has yet to be achieved. In this study, we present an AVF that autonomously closes and reopens, utilizing novel thermo-responsive UV-curable shape memory materials for soft robotic systems. The life-sized thermo-responsive AVF exhibits closing and reopening motions triggered in a naturally occurring temperature range. The doubly curved trap lobes, built from shape memory polymers, close at 38°C, while reopening initiates around 45°C, employing shape memory elastomer strips as antagonistic actuators to facilitate lobe reopening. This work represents the first demonstration of thermo-responsive closing and reopening in an AVF with programmed sequential motion in response to increasing temperature. This approach marks the next step toward autonomously bidirectional moving soft machines/robots.

## 1       Introduction

Nature has inspired many technological advancements, while animals provide a deep well of inspiration, the field of soft machines and soft robots has recently steered its focus on the inspiration that plants provide. Specifically, how their principles behind energy efficiency, sustainability, growth and motion, can be translated into technology [1]. A fascinating model for plant motions is *Dionaea muscipula*, commonly known as the Venus flytrap. Its leaves evolved into specialized snap traps consisting of two bilaterally symmetric, doubly curved trapezoidal lobes connected by a midrib [2–5]. The traps typically close within 100-500 ms in response to two consecutive stimuli. In contrast trap reopening typically is much slower depending on an unsuccessful (hours) or successful prey capture (days).

In recent years the fast snap buckling closure motion of *D. muscipula* has inspired and was transferred into a variety of artificial Venus Flytraps (AVFs) systems [6–12]. Utilizing different actuator principles, like pneumatics, magnetic or shape memory alloys, these systems have been engineered to achieve closure motions with snap closures and curvature inversions at speeds comparable to the biological role model (100-500 ms) [6]. However, the reopening of the trap after closure has not been thoroughly examined in any of these examples. In these technical systems, reopening is often a byproduct of stopping the actuation.

In nature, the process of trap reopening is a bit more elaborate. Until 2022, trap reopening was considered a "slow, smooth process driven by hydraulics and cell growth" [13]. Durak et al. [13] demonstrated that trap reopening is also contingent on trap morphology. Utilizing three-dimensional digital image correlation (3D-DIC), as well as morphological and mechanical investigations, they conclude that differences in trap reopening stem from a combination of size, length-to-height ratio and slenderness of the individual traps. Biological and physiological factors also contribute to changes in the observed behavior. Durak et al. found that under specific conditions, in addition to the typically observed gradual reopening, slender and large traps (approximately 4.5 cm in length) can undergo a smooth bending during the opening process, that subsequently leads to a rapid snap-through of the lobes.

We propose a slender trap design and antagonistic motion of shape memory materials with different switching temperatures to achieve a snap closure and reopening of the trap. In order to create an AVF demonstrator that closes and reopens in sequence, the demonstrator needs to be actuated sequentially. This can be achieved with a multilayered design as highlighted by Tauber et al. [12]. They utilized different pre-stretch ratios and a complex bonding process to create a multilayered trap lobe with a contracting shape memory polymer inner side, bonded to a silicone membrane, in turn bonded to an outer layer with a high temperature expansion coefficient. The system closed at a high switching temperature of 70°C and only reopened marginally after continuous heating. Therefore, in this study, we tackle this challenge by combining two different types of thermo-responsive materials shape memory polymers and shape memory elastomers to create a thermo-responsive autonomous system functioning in the range of natural temperatures. Enabling functionality in naturally occurring temperature ranges could increase the autonomy of soft machines and robotic systems.

## 2      Shape memory materials

In order to enable an AVF to close and reopen in response to a rise in temperature within a natural occurring temperature range, we propose the use of two different shape memory materials with a one-way shape memory effect. These materials could work antagonistically, so that when combined, they create a pseudo two-way shape memory effect.

Shape memory materials, like the shape memory polymers (SMP) and elastomers (SME), can be deformed into temporary shapes and recover their original shape upon heating beyond a specific threshold temperature. During the programming process (**Fig. 1**) [14], the material is first heated above a certain threshold temperature, also referred to as the switching temperature $T_{sw}$. Above this temperature, which can be the glass transition ($T_g$) or the melting point ($T_m$), the polymer can be deformed into any shape, by applying deformational stress (*Deformation*). The material is then cooled to below the transition point, while maintaining the stress (*Cooling/Fixing*). Subsequently, the stress is removed. Depending on the material's shape fixity (ability to retain the temporary state - thus store strain energy), the material may deform slightly during the unloading process (*Unloading*). By reheating the SMP above the $T_{sw}$ the material returns to its original permanent shape (*Recovery*) completing the shape memory cycle. **Fig. 1** displays an exemplary shape memory cycle of an SMP with a one-way shape memory effect. Such materials require manual reprogramming after each shape memory cycle [14, 15]. The glass transition temperature ($T_g$) refers to the temperature at which a material experiences a phase transition from a rigid, glassy state to a more flexible, rubbery state. Unlike a phase change like melting, this transition represents a change in the physical properties of the material. Below the $T_g$, molecular motion is restricted, resulting in a rigid and glassy state. Conversely, when temperatures rise above the $T_g$, the material transitions to a softer and more flexible condition, as such the shape memory materials $T_g$ is chosen as the planned systems $T_{sw}$ [14,15].

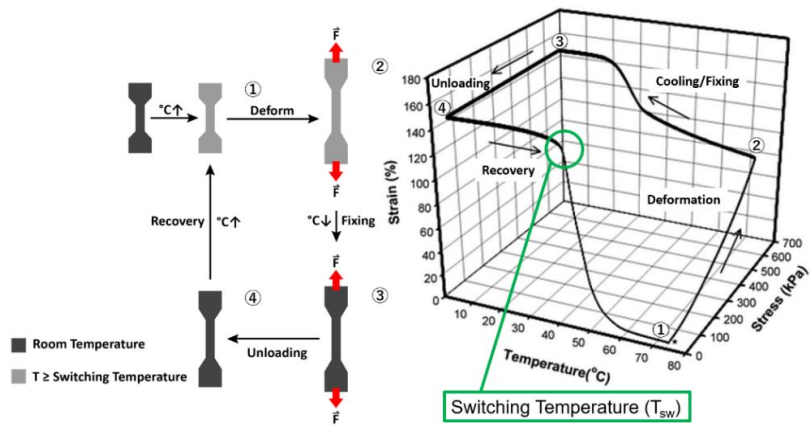

**Fig. 1.** Exemplary shape memory cycle of a shape memory polymer. (Reproduced and modified from Ref. [15] with permission from the Royal Society of Chemistry).

Achieving sequential actuation in an AVF requires two types of shape memory materials: a stiffer material as the base, and a more elastic antagonist. The first material must have a high stretchability above its $T_{sw}$ for form stability, while the second should be deformable below its $T_{sw}$ and should generate sufficient force to restretch the first material. We aim to use a highly stretchable shape memory polymer (SMP) as the first material and a shape memory elastomer (SME) with high elasticity at room temperature as the second. Each layer must have different switching temperatures to facilitate motion, causing the AVF lobes to open and close [14, 16].

For the first SMP layer, an aliphatic urethane diacrylate (AUD) based crosslinking system was used as a basis due to its compatibility with UV-curing methods. Genomer 4230 (Genomer 4230 Rahn AG) was used as an AUD component with a concentration of 5wt% diluted in isobornyl acrylate (IBOA). Diphenyl(2,4,6-trimethylbenzoyl) phosphine oxide (TPO) was added as a photoinitiator. In preliminary tests, different additive compounds were tested for their influence on the stability of the final material and its shape memory properties. Lauric acid (LA) was identified as the best candidate due to its material properties (elasticity, shape fixity, switching temperature). Three different concentrations were further investigated (20wt%, 30wt% and 40wt%) to determine how they affect the switching temperature, elasticity, strain recovery and fixity. A SMP with a 20wt% concentration of LA proved to have a suitable switching temperature of 38.1°C. It also had the highest elasticity and strain recovery of the three.

For the second material, we used an SME material based on the UV curable resin developed and provided by our colleagues in the NeptunLab (University of Freiburg) [17]. This polyethylene glycol-based material with a long-chain polyethylene glycol dimethacrylate (PEGDMA) as crosslinker and 2-carboxyethyl acrylate (CEA) as monomer was considered suitable as it showed high elasticity (~400%), good shape memory properties (Shape fixity ratio $R_f > 97\%$, shape recovery ratio $R_r > 99\%$). In preliminary tests, different crosslinker ratios of 25wt% and 40wt% were investigated on how these affect the materials $T_{sw}$. The aim was to ensure that the SME had a significantly different $T_{sw}$ from the SMP to achieve effective sequential actuation. The properties of these materials were expected to enable the demonstrator to actuate via snap buckling mechanisms. Tensile testing in a temperature chamber without external stress was conducted to measure the recovery force of L20, SME25, and SME40. The results showed that SME25 and SME40 produced consistent forces, with an increase starting between 20-40°C, while L20 exhibited diverse starting points (**Fig. 2**). No significant force change occurred beyond 60°C for all materials, indicating full polymer chain retraction. L20 emerged as the optimal candidate for the opening motion due to its actuation temperature and force characteristics, while SME25 was chosen for the closing motion as its $T_{sw}$ was significantly different to L20.

Both materials were used as UV-curable resin in mold casting processes with flexible PDMS molds to prepare the sample for material characterization as well as for the AVF demonstrators. The focus on UV curability was intended to enable the mass production of AVFs using light-based additive manufacturing techniques – such as stereolithography (SLA), digital light processing (DLP), and volumetric 3D printing – if the material tests confirmed suitability as an actuator material and the AVF demonstrator achieved sequential motions.

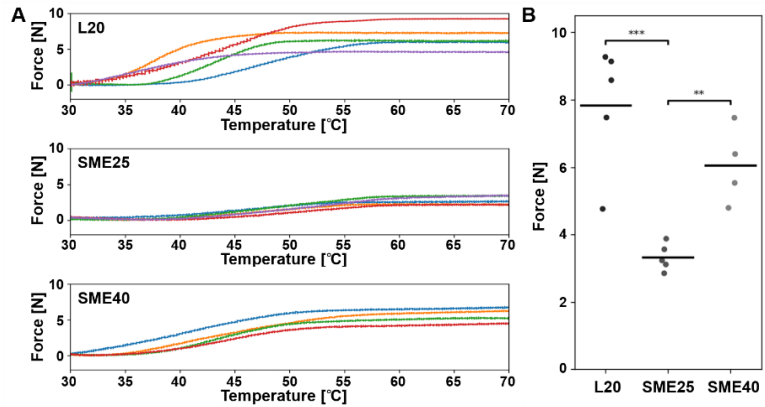

**Fig. 2.** Shape recovery characterization of the shape memory materials. A) Exemplary shape recovery force experiments plotted over a temperature range of 40°C for L20, SME25 and SME40. B) Maximum force generated through shape recovery of L20 (n=6), SME25 (n=5) and SME40 (n=4). The black horizontal bars represent the mean value. L20 developed higher forces than both SME25 and SME40. Significant differences in force could be observed between L20 and SME25 (p-value <= 0.001) and between SME25 and SME40 (p-value <= 0.01).

## 3  Artificial Venus Flytrap demonstrator

### 3.1  Concept of a thermo-responsive closing and reopening AVF

Resulting from the material analysis, we developed a conceptual design for the closing and reopening AVF utilizing the two antagonistically functioning shape memory materials. **Fig. 3** depicts a conceptual sequence of the AVFs motion cycle. First a rise in temperature to 38°C triggers the shape recovery of the SMP, which snaps the AVF closed ($T_{sw}$ of SMP reached). Simultaneously, the SME material stretches, as in theory its elasticity increases with an increase temperature. When the temperature increases beyond a certain threshold, the $T_{sw}$ of the SME is reached, triggering its shape recovery, then the SME contracts, reopening the AVF and re-stretching the SMP.

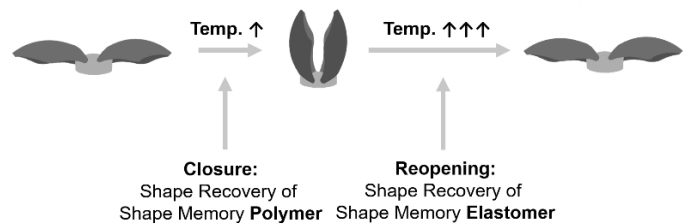

**Fig. 3.** Concept of the thermo-responsive AVF utilizing shape memory materials for closing and opening. Multi-material lobes are intended with an inner shape memory polymer layer (light grey) and an outer shape memory elastomer layer (dark grey).

### 3.2    Mono-material shape memory AVF demonstrator

In order to realize the intended motion sequence, it first had to be demonstrated that the two shape memory materials were suitable for producing a snapping AVF. Therefore, mono-material demonstrators were produced and their reaction to a temperature increase above the $T_{sw}$ of the material was investigated. In accordance with the requirements of Durak et al. [13], long and slender AVF lobes were designed in different parameter combinations, by varying the horizontal lobe length and lobe thickness. The 3D-scanned lobe model by Wang et al. [9] was used as a base model for the flytrap lobe shape and were slightly adjusted by adding curvature to the lobes. A positive gaussian curvature (K) was chosen, which can be described by the product of the two principal curvatures ($\kappa 1$ and $\kappa 2$), with principal radii R1 and R2 of 40mm and 110mm, respectively. Preliminary tests with individual lobes showed that these radii led to the curvature inversion of the lobe. For the fabrication of the single material demonstrators, two of the lobes were connected at the bottom via a midrib from the same material, forming an artificial Venus flytrap trap. For the single material demonstrators, differently sized clear PDMS molds were created. The shape memory resins were then liquified by heating and then poured into the molds. The resin was cured inside the PDMS mold by irradiating it with 365nm UV light and subsequently further hardened at room temperature overnight. Once fully cured, the lobes were programmed by inverting their curvature from a concave to a convex shape by bending.

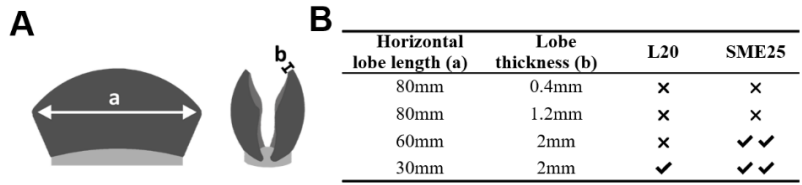

**Fig. 4.** Mono-material AVF parameters. A) Sketch of the single material actuator design with the horizontal length 'a' and the lobe thickness 'b' and the implemented curvature. B) Lobe dimensions investigated for the closure of the AVF. Right columns show the motion analysis results of observed lobe closure after heating of different lobes with varying thickness to horizontal lobe size ratios with both shape memory materials. '✓' represents a successful closure, while '✗' marks failed closing. '✓✓' represents successful closure coupled with snap buckling.

To characterize the closing motion of the AVF demonstrators, video analysis and motion tracking were performed for which all specimens were placed in a climate test chamber (CTC 256, Memmert GmbH, Schwabach, Germany) and heated from 20°C up to 70°C, with a temperature increase of 1C/min. This process was captured with a SLR-camera (25 fps) for further analysis with the motion tracking software Kinovea (Version 0.9.5). This software was used to track the bending angles of the actuators and to track the lobe ends of the demonstrators to measure their range of motion. To trace the motion in more detail, we attached either black dots on the lobe sides or small white spheres with black dots to the specimen. The results of the characterization highlighted that thicker lobes (2 mm) with at least 30 mm length were capable of closure (**Fig. 4**

B). However, only SME25 showed a snap buckling during closure at around 45 °C (**Fig. 4** B, **Fig. 5** A-C), indicating a possible rapid release of stored elastic energy in the system. In contrast to the SME25 AVF, the L20 AVF closure motion started around 30 °C and was finished at 40 °C with main motion around 38°C $T_{sw}$ of L20 (**Fig. 5** D). Unlike SM25, the L20 AVF showed no observable snap buckling during the curvature inversion as it smoothly closed (**Fig. 5** E, F). The SM25 AVFs closure motion with snapping was too violent (jumped during snap closure) to capture video motion tracking as the jump and snap happened within 1 frame. The characterization of the mono-material demonstrators showed that both shape memory materials, SME25 and L20, could be used to create a closing system. For fabricating the closing and reopening system, L20 was selected for the closing lobes and SM25 for the reopening mechanism due to their $T_{SW}$. While L20 does not generate a snapping motion, the difference in $T_{sw}$ was prioritized, for the moment, to develop a sequential closing and reopening system.

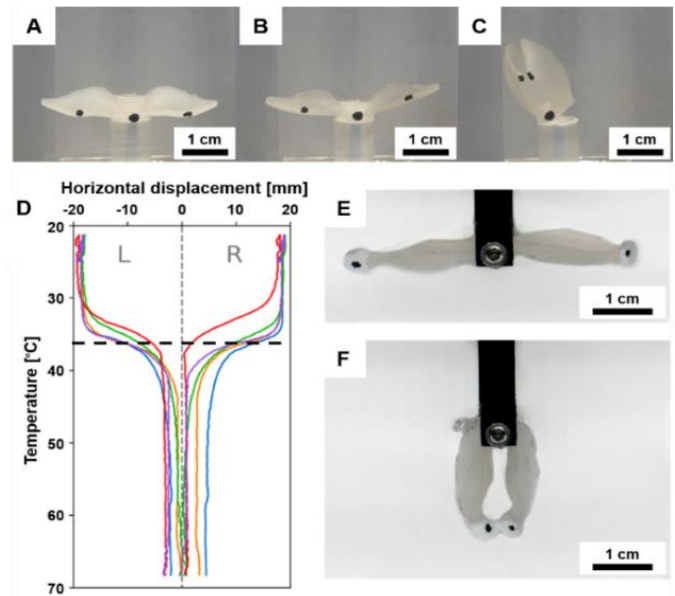

**Fig. 5.** Mono-material AVF characterization. A-C: Mono-material actuator made of SME25 during testing inside the climate chamber with a lobe thickness of 2 mm and a horizontal lobe length of 60 mm. At room temperature the AVF is in the initial open state at room temperature (A). A first motion is notable before reaching the $T_{sw}$ (B). Approximately 1 second after reaching $T_{sw}$ the AVF is closed after snap buckling occurred (C). In comparison, tracked motions of the L20 AVFs (D): the lines represent the tips of each actuator lobes replicates (n=5). The tips are considered close if the lines reach the 0 mm reference line (light grey dotted line) in the middle. The $T_{sw}$ of 38 °C is marked as a dashed line. The AVF switches its curvature in a smooth motion from the initial opened state (E) to the closed stated (F) no snap buckling was visible although the lobe curvature inversed. The black dots on the tips of the lobe were used to track the movement of the actuator.

### 3.3    Closing and reopening AVF demonstrator

Fabricating and characterizing a closing and reopening AVF demonstrator is a multi-step process. It requires: (1) mold casting the lobes from different shape memory materials, (2) programming desired shape recovery motion, (3) bonding both materials, and (4) testing the motion. Unfortunately, the preliminary tested approach to use full lobe sized actuators bonded together did not achieve functional demonstrators. Either the lobes did not move or only the SME25 triggered. From these results it was deduced that the L20 does not develop enough force to stretch the SME25 material at L20 $T_{sw}$. To overcome this problem, we investigated other approaches to reduce the necessary forces. Preliminary experiments showed that SME25 programmed with 40% strain could be used as small linear contraction actuators. Programming with stretching or straining was shown by Tauber et al. [12] as a viable method to increase contraction rate in SMP materials. To achieve a functioning AVF demonstrator that can move bidirectionally, we chose a combination of stretched SME25 strips with preprogrammed L20 lobes (**Fig. 6** A). The 2 mm thick L20 lobes were programmed by bending the doubly curved lobes, thus creating curvature inverted lobes. Afterwards, thin SME25 strips, strained at 40%, were attached with glue (Loctite 454) in three different configurations as 'single strand', 'cross shape' and 'diamond shape' (**Fig. 6** B). For the purpose of characterization, several specimens of each configuration were produced and heated above their $T_{sw}$ once each and the behavior recorded.

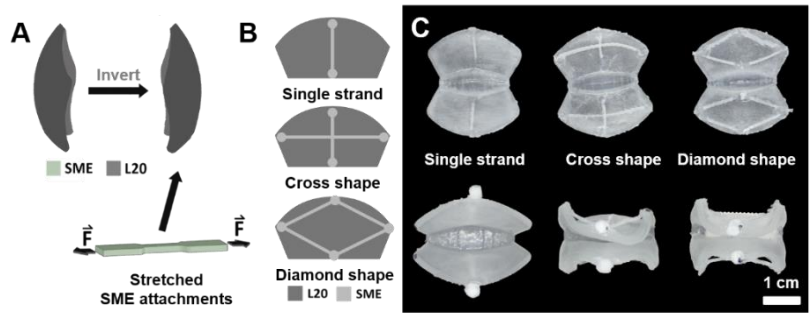

**Fig. 6.** Production of bidirectionally moving AVF demonstrators. A) The L20 lobes were programmed with an inverted curvature. Preprogrammed strips of SM25 with 40% strain were attached onto the lobes in three different configurations 'single strand', 'cross shape' and 'diamond shape' (B). C) The three demonstrator types from left to right in their programmed open state (top) and their closed and reopened state after actuation (bottom): 'single strand', 'cross shape', 'diamond shape'. For kinematic tracking, white spheres were attached to the lobe tips.

The lobe size and thickness of the single material unidirectional actuator were used as a basis for the demonstrator. The SME strands had a cross section of 0,6mm x 0,6mm, as the thin design enabled a reduction of resistance that was observed in preliminary tests. When triggered, we predicted that the L20 lobes would invert their curvature, thus closing the AVF and conversely triggering the SM25 strands would act as contraction actuators reopening the lobes. The following section addresses the predicted motions based on the chosen designs.

In the 'single strand' design, the strand was arranged colinearly with the recovery motion of the lobe. This arrangement could have reduced the closure, but also could have improved the reopening as the main motion axis would be pulled back. In the 'cross shape' design, the additional strand was expected to improve the recovery of the original double-curved shape by introducing an additional actuation axis. However, this configuration could also have imposed additional restraining forces on the L20 shape recovery. The third 'diamond shaped' design was inspired by the actuator of Masselter et al. [18] and aimed to mitigate the resistance stemming from the SME strands below their $T_{sw}$ by arranging them in a 'diamond shaped' geometry. This configuration was predicted to enable the shape recovery force to act in an angular, lever-like manner on the strands, deviating from the direct, colinear orientation. The 'diamond shape' was designed with a width-to-height ratio of 1:2, emulating the strains observed in a biological Venus flytrap [2, 5, 13]. All proposed designs were producible (**Fig. 6** C) and reacted to an increase in temperature over 10 min with sequential closing at $T_{sw}$ of L20 and opening at $T_{sw}$ of SME25 (**Fig. 7** A). **Fig. 7** (A) depicts the motion tracking as the horizontal displacement of the left and right lobe tips of the demonstrator over increasing temperature. Within this figure, two events are observable. First, both lobes approach the 0 mm reference line at around 40°C, which is associated with the closing motion. Reaching the 0 mm reference line equals a fully closed system. Second, this is followed by an increase in distance of the lobe tips, marking the reopening of the lobes beyond 50°C. This pattern is observable in all three designs. In order to analyze the sequential motion sequences more closely, the relative motion of the closing and reopening phases was considered separately and the total range of motion of the demonstrators was determined (**Fig. 7** B). All values indicated are relative to the maximum possible values. A relative movement of 100% for closure therefore represents that the demonstrator is capable of fully closing the lobes, which equals the shape before programming. On the other hand, a full reopening indicates that the demonstrator is capable of returning to the (fully) open shape it had after programming. The overall range of motion (ROM) represents, as a value, the complete bidirectional movement. The ROM is relative to the maximum moveable distance in total and is calculated as the mean value of closure and reopening. The motion analysis of the three demonstrator types with the different SM25 strand configurations showed that closing and reopening is achievable with antagonistically working shape memory actuators (**Fig. 6C**, **Fig. 7** B).

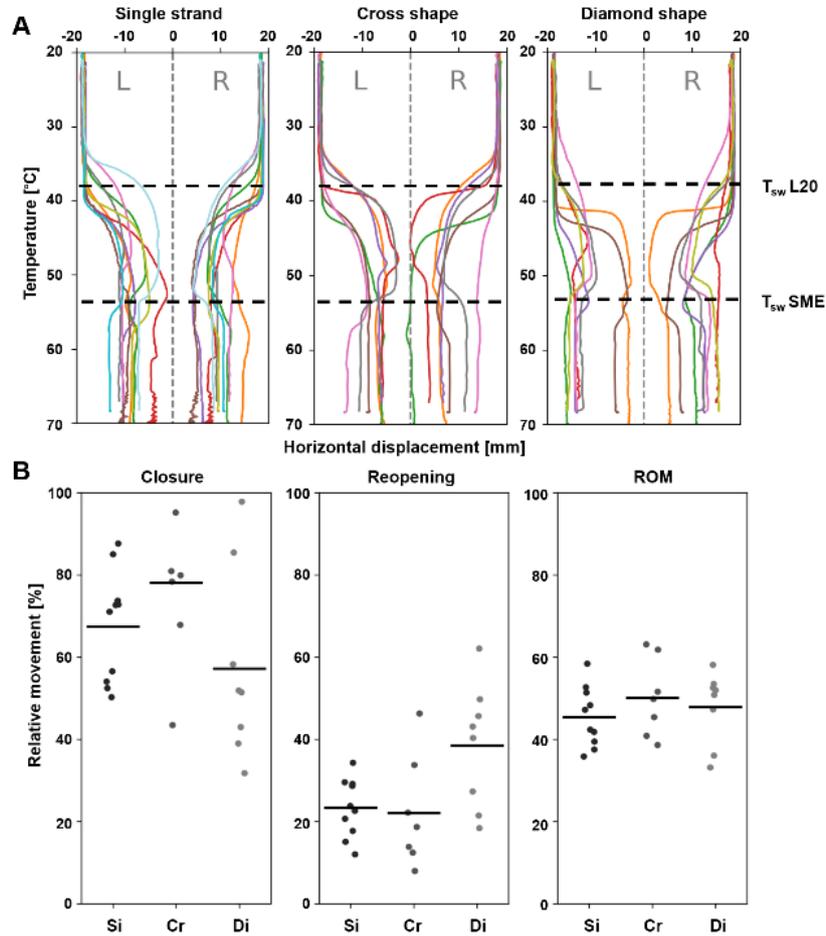

**Fig. 7.** Motion analysis of the AVF demonstrator variants: single strand (Si) (n=10), cross shaped (Cr) (n=7), diamond shaped (Di) (n=8). A) Horizontal displacement of the AVF specimens, using three different geometries for the alignment of the SME strands. The different colored lines represent the position of the left and right outer lobe tips of each AVF configuration replicates in correspondence to the temperature. B) Comparison of the relative movement of the AVFs, in three subcategories: closure, reopening and range of motion (ROM). The 'diamond shape' configuration achieved the highest relative closing and reopening motions. The ROM is similar between all three tested configurations.

While no significant difference was found between the 'cross shaped' and 'single strand' types, notable differences were observed to the 'diamond shaped' type. The 'diamond shape', designed to reduce resistance during closure through angular arrangement of SME strands and an open lobe center, showed improved lobe reopening. This is likely due to its increased number of strands, despite the non-parallel orientation of the strands with respect to the reopening motion. This configuration potentially enables more uniform force distribution, resulting in a wider lobe reopening. The initial prediction that horizontally applied force in the 'cross shaped' actuator contributed to wider reopening proved to be wrong. As such, two conclusions arise: The arrangement of SME strands impacts curvature during closure and reopening, and an increased number of strands as in the 'diamond shaped' type is a key factor in reopening. Both conclusions, however, require further validation. Comparing the mono-material AVFs to the reopening demonstrators shows that none achieved 100% lobe closure, indicating that SME strand attachment hinders closure and reopening range. The use of glue (Loctite 454) may also contribute to reduced elasticity at attachment points. The 'cross shaped' and 'diamond shaped' types likely face more closure difficulties due to their higher number of strands and anchor points, yet no significant differences were found. To improve closure, the strands could be attached with some slack to allow freer lobe movement. Utilizing the slender trap design facilitated the closing and reopening of the designed traps. This design's efficacy is underlined by a direct comparison to a previous AVF version in which a bigger doubled curved sheet design only achieved marginal reopening [12]. Some features from the biological model could not be adapted, such as the rapid actuation time of the Venus flytrap, which closes in fractions of a second compared to the 10-minute average of the AVFs in this study. This rapid closure is attributed to a unique snapping mechanism, which was intended to be replicated using elastic materials capable of storing energy. However, snap buckling was not achievable due to the properties of L20 flexibility. Furthermore, current demonstrators cannot fully reset, as L20 and SME25 are unable to generate enough force to stretch the opposing set. By system design the measurement of forces of the individual materials and the opening and closing forces was not possible. An estimate based on the material test data indicate that the LM20 should generate enough to stretch force the thin SME strands. As such, a cyclic reset in the current designs is not possible and a system reset is only possible by replacing SME strands, hindering autonomy. Comparison with other recent artificial Venus flytraps suggests that, while improvements in closing time are needed, the development of a competitive demonstrator has been successful, creating for the first time a thermo-responsive closing and reopening AVF based on shape memory polymers. With the proposed elements, snap buckling and enhanced closing times via direct heating could be possible. Despite the implementation of fast reversible bidirectional movement in various AVF demonstrators, thermally actuated versions remain largely unexploited. This study comprehensively examined the viability of a closing and reopening thermo-responsive AVF demonstrator with a more natural slender design that is capable of actuation in a natural temperature range.

## 4     Conclusion

This work successfully demonstrated the development of an artificial bidirectionally moving Venus flytrap demonstrator using thermal actuation. Unlike many existing systems, which employ air- or electricity-based actuation, this approach created a passively acting soft systems that respond to environmental changes like temperature or sunlight intensity. However, the high flexibility of L20 led to some limitations, namely, it exhibited excessive softness at elevated temperatures and impeded the snap buckling behavior. Consequently, this complicated the development of a sequentially moving AVF. By employing antagonistically working SME25 strip actuators, a sequential motion was achieved. Although a cast molding process with UV-curing was used, the SMPs developed could be used for 3D printing. If novel multi material light-based 3D printing techniques, like volumetric printing, are employed [19, 20], constructing an entirely 3D-printed multi-material demonstrator could be feasible. This would allow the fabrication of detailed structures, such as thin mesh layers, promoting uniform force distribution and improving curvature inversion. Additionally, this approach could eliminate the need for adhesives and enhance actuation, thus enabling snap buckling. Furthermore, increasing the reaction and actuation speed of these systems requires fast heat transfer. Additives like metal particles or photothermal agents could enable the systems to react faster to heat and might even allow for photothermal responsiveness. This work demonstrated the successful development of an AVF using thermo-responsive shape memory materials, marking a step toward the feasibility of bidirectional moving autonomous soft robots for a potential use in environmental monitoring.

**Acknowledgments.** Funded by the Deutsche Forschungsgemeinschaft (DFG, German Research Foundation) under Germany's Excellence Strategy – EXC-2193/1 – 390951807. We thank Laura Mahoney from the livMatS Writer studio for proofreading and improving the manuscript.